\documentclass{article}

\PassOptionsToPackage{numbers,sort&compress}{natbib}

\usepackage[dblblindworkshop, final]{neurips_2025}

\workshoptitle{Embodied and Safe-Assured Robotic Systems}

\usepackage[utf8]{inputenc} %
\usepackage[T1]{fontenc}    %
\usepackage{hyperref}       %
\usepackage{url}            %
\usepackage{amsfonts}       %
\usepackage{nicefrac}       %
\usepackage{microtype}      %
\usepackage{xcolor}         %
\usepackage{dsfont}
\usepackage{makecell}

\usepackage{booktabs,threeparttable,siunitx,xcolor}
\usepackage{graphicx}
\usepackage{multirow}
\usepackage{algorithm}
\usepackage{algorithmic}
\usepackage{natbib}
\usepackage{caption}
\usepackage{subcaption} %
\DeclareCaptionLabelFormat{bold}{\textbf{#1 #2}}
\usepackage{wrapfig}
\usepackage{amsmath}

\newcommand{\Hnorm}[1]{\frac{#1}{\ln |\mathcal H|}}

\usepackage{newfloat}
\usepackage{listings}

\usepackage{siunitx}
\usepackage{titlesec}
\titlespacing*{\section}{0pt}{0.05ex plus 0.1ex minus 0.05ex}{0.3ex}
\titlespacing*{\subsection}{0pt}{0.3ex plus 0.2ex minus 0.1ex}{0.2ex}
\titlespacing*{\subsubsection}{0pt}{0.3ex plus 0.1ex minus 0.1ex}{0.1ex}
\newfloat{listing}{tb}{lst}{}
\floatname{listing}{Listing}

\newfloat{listing}{tb}{lst}{}
\floatname{listing}{Listing}

\lstdefinelanguage{PDDL}{
  morekeywords={define,domain,and,or,not,when,forall,exists,increase,decrease,assign},
  otherkeywords={:requirements,:types,:predicates,:functions,
    :action,:parameters,:precondition,:effect,:problem,:objects,:init,:goal,
    :strips,:typing,:fluents,:equality,:adl},
  sensitive=false,
  morecomment=[l]{;},
  morestring=[b]",
  alsoletter={:?-}
}
\title{Gated Uncertainty-Aware Runtime Dual Invariants for Neural Signal-Controlled Robotics}

\author{%
  Tasha Kim\\
  Oxford Robotics Institute (ORI)\\
  Department of Engineering Science\\
  University of Oxford\\
  \texttt{tashakim@eng.ox.ac.uk} \\
  \And
  Oiwi Parker Jones \\
  Oxford Robotics Institute (ORI)\\
  Department of Engineering Science \\
  University of Oxford \\
  \texttt{oiwi.parkerjones@eng.ox.ac.uk} \\
}

\begin{document}

\maketitle

\begin{abstract}
Safety-critical assistive systems that directly decode user intent from neural signals require rigorous guarantees of reliability and trust. We present GUARDIAN (Gated Uncertainty-Aware Runtime Dual Invariants), a framework for real-time neuro-symbolic verification for neural signal-controlled robotics. GUARDIAN enforces both logical safety and physiological trust by coupling confidence-calibrated brain signal decoding with symbolic goal grounding and dual-layer runtime monitoring. On the \textsc{BNCI2014} motor imagery electroencephalogram (EEG) dataset with 9 subjects and 5,184 trials, the system performs at a high safety rate of 94--97\% even with lightweight decoder architectures with low test accuracies (27--46\%) and high ECE confidence miscalibration (0.22--0.41). We demonstrate $\approx$1.7x correct interventions in simulated noise testing versus at baseline. The monitor operates at 100Hz and sub-millisecond decision latency, making it practically viable for closed-loop neural signal-based systems. Across 21 ablation results, GUARDIAN exhibits a graduated response to signal degradation, and produces auditable traces from intent, plan to action, helping to link neural evidence to verifiable robot action.

\end{abstract}

\section{Introduction}
Neural signal-controlled robots have significant potential to improve accessibility for individuals with limited mobility but they also introduce important safety risks\citep{Stolzle2024}. Maintaining runtime accuracy and implementing reliable intervention mechanisms are paramount in closed-loop systems to ensure user safety\citep{astmf3269,cofer2020taxiing,huang2014rosrv}. Recent closed-loop EEG-based assistive systems~\citep{Zhang2025EEGBCI, kim2024noir2} and AI-enabled brain-computer interfaces (BCIs) ~\citep{Lee2025} show encouraging progress, but remain vulnerable to ambiguous user intent, a known challenge in shared autonomy and teleoperation\citep{dragan2013policy,javdani2018ijrr,javdani2015rss}. They also suffer signal degradation during long-horizon tasks\citep{jayaram2016transfer,spuler2012covariateshift,muller2008singletrial,rashid2020status} and lack formal or interpretable safety mechanisms like runtime assurance\citep{astmf3269,huang2014rosrv}, or shielding\citep{alshiekh2018shielding}. %
We propose GUARDIAN, a physiological runtime verification architecture that provides an auditable, explainable layer between neural decoding and execution. %
GUARDIAN acts as a safety gate that complements neurally-controlled systems, producing tracing and intervention tooling without disrupting real-time operations.

\textbf{Design principle and goals.} GUARDIAN prioritizes the following principles: (a) conservative safety (\emph{halting} over execution when neural evidence is weak or contradictory), (b) interpretability (traceable action chains from raw brain signal$\to$intent$\to$plan$\to$action compatible with common symbolic planning toolchains like PDDL~\citep{ghallab2004pddl}), (c) modularity (wraps any decoder with $<1$ms overhead), and (d) tunability (operators can set confidence thresholds informed by calibration analysis). %

\textbf{Threat model and scope.} Our system operates with non-invasive neural signals in assistive robot domains, oftentimes challenging due to noise interference, non-stationarity--within-session accuracy drop and between-session variance, neural artifacts, and subject fatigue\citep{rashid2020status,bigdely2015prep}. We address confidence miscalibration, where decoders show high confidence despite poor accuracy\citep{guo2017calibration,minderer2021revisiting}. When there are no violations, our monitor gates actuation and operates under standard action safety protocols\citep{iso10218-1,iso10218-2,iso15066}. In assistive robotics, hardware can pose physical risks for users when there is uncontrolled device activation or delayed shutdown\citep{iso15066,astmf3269}. Attacks or adversary threats are outside our scope. %

\textbf{Contributions.} We make the following contributions: (I) a safety-centric framework achieving high safety rates under severe signal degradation, %
(II) a verifiable neuro-symbolic pipeline that formally links EEG distributions to symbolic goals, and (III) a practical lightweight architecture with minute computational overhead enabling deployable and trustworthy neural signal-based control.

\vspace{1mm}

\section{GUARDIAN: Formalization and Algorithm}

\begin{wrapfigure}{l}{0.55\textwidth} %
  \vspace{-8pt}
\centering
\includegraphics[width=0.55\textwidth]{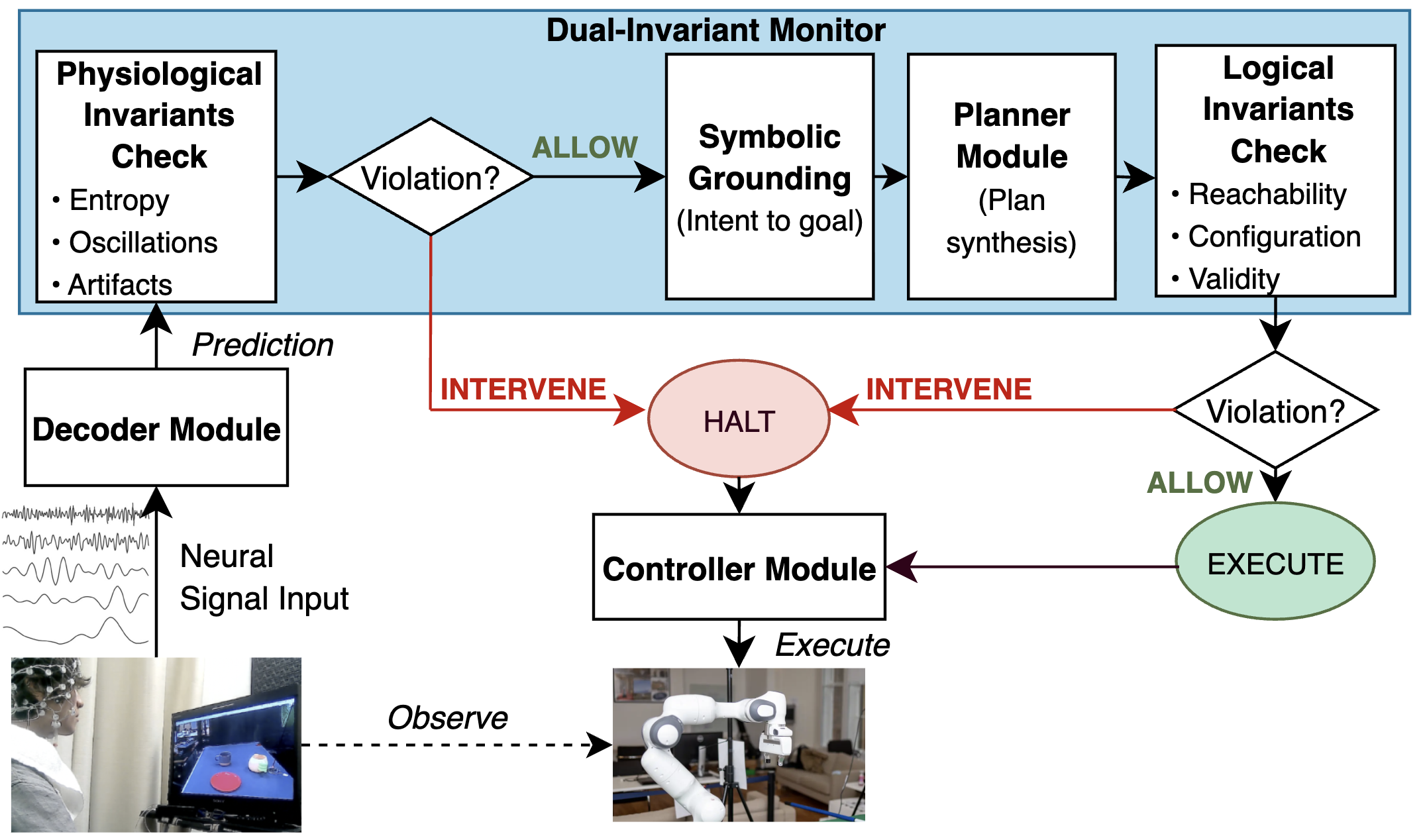}
\caption{\textbf{System architecture overview.
}}
\label{fig:sys}
\end{wrapfigure}

Let $x_t\in\mathbb{R}^{C\times W}$ denote the EEG window at time $t$ ($C$ channels, $W$ samples). A decoder $f_\theta$ outputs a categorical posterior $p_t\in\Delta^3$ over the action set $\mathcal{H}=\{\textsc{grasp}, \textsc{release}, \textsc{move\_to}, \textsc{rotate}\}$ corresponding to left hand, right hand, feet, and tongue motor imagery (MI), respectively. These actions enable natural control (positioning, grasping, releasing, orienting) for EEG-based manipulation. We treat $p_t$ as a $4$-dimensional probability vector over $\mathcal{H}$, i.e., $p_t \in \Delta^{3} \subset \mathbb{R}^4$ (3-simplex). Let $\mathbf{u}$ denote the uniform distribution on $\mathcal{H}$, i.e., $\mathbf{u}(h)=\tfrac{1}{|\mathcal{H}|}=\tfrac14$ for all $h\in\mathcal{H}$. Unless stated otherwise, $\log$ denotes the natural logarithm and $0\log 0 := 0$.
\vspace{-2mm}

\paragraph{Calibration-aware safety.}
To mitigate the effects of miscalibration, we construct a calibrated intent distribution by convexly mixing with the uniform prior \(\tilde p_t \;=\; \alpha_m p_t + (1-\alpha_m)\,\mathbf u, \alpha_m \in [0.5, 0.8],\) where \(\mathbf u\) is the uniform distribution on \(\mathcal H\). Since this is a convex combination, \(\tilde p_t \in \Delta^{3}\). Epistemic uncertainty is quantified using the normalized Shannon entropy
\begingroup
\setlength{\abovedisplayskip}{3pt}
\setlength{\belowdisplayskip}{3pt}
\[
H(\tilde p_t) \;=\; -\sum_{h\in\mathcal H}\tilde p_t(h)\,\log \tilde p_t(h),
\qquad
\hat H(\tilde p_t) \;=\; \Hnorm{H(\tilde p_t)} \in [0,1].
\]
\endgroup
Rapid intent flips are detected by computing the oscillation index
\begingroup
\setlength{\abovedisplayskip}{3pt}
\setlength{\belowdisplayskip}{3pt}
\[
\Omega_t \;=\; \frac{1}{K-1}\sum_{k=t-K+1}^{t-1}
\mathds{1}\!\left\{
\operatorname*{argmax}_{h\in\mathcal H}\tilde p_{k}(h)\,
\neq \,
\operatorname*{argmax}_{h\in\mathcal H}\tilde p_{k+1}(h)
\right\},
\]
\endgroup
over the last \(K\) frames (\(K{=}10\) at \(100\,\mathrm{Hz}\), \(K\!\ge\!2\)). To obtain a scalar artifact score comparable to a threshold, the per-channel band-limited RMS energies in \([20,45]\,\mathrm{Hz}\) are z-scored against a subject baseline and aggregated as the mean across channels, 
\begingroup
\setlength{\abovedisplayskip}{1pt}
\setlength{\belowdisplayskip}{1pt}
\[A_t \;=\; \frac{1}{C \; }\sum_{c=1}^C \; \mathrm{zRMS}_t^{(c)}.\]
\endgroup
Here, \(\mathrm{zRMS}_t^{(c)}\) denotes the z-scored root-mean-square amplitude of channel \(c\) within the \([20,45]\,\mathrm{Hz}\) band (dimensionless after z-scoring). Mean aggregation was empirically found to be robust. Note that max-based aggregation may also be used with an appropriately tuned threshold \(\tau_A\).

\textbf{Invariant types.} Let $\tau=(\tau_H,\tau_A,\tau_\Omega)$ denote the physiological thresholds. We define \emph{physiological invariants} $\Psi=\{\psi_1,\psi_2,\psi_3\}$ and \emph{logical invariants} $\Phi=\{\phi_1,\phi_2,\phi_3\}$ as
$\psi_1: \hat H(\tilde{p}_t) < \tau_H$, 
$\psi_2: A_t < \tau_A$, 
$\psi_3: \Omega_t < \tau_\Omega$, 
$\phi_1:$ objects reachable, 
$\phi_2:$ safe configurations, 
$\phi_3:$ valid transitions.

\textbf{Decoder architectures.}
We evaluate four representative neural decoder architectures: EEGNet~\citep{lawhern2018eegnet}, Riemannian covariance model~\citep{barachant2012riemann}, lightweight CNN~\citep{schirrmeister2017deep} and an interpretable feature model (RealIntent) each spanning 12k--45k parameters. See full architecture and training details in App.~\ref{app:decoders}.

\textbf{Dual safety monitoring.}
The dual-layer monitor (See Alg.~\ref{alg:monitor}) checks physiological invariants $\Psi$ (lines 4--6), grounds intent to a symbolic goal (line 7), synthesizes it into a plan (line 8), and verifies logical invariants $\Phi$ (lines 9-11). Any violation triggers an intervention, which halts the execution and maintains the current safe state \textsc{IDLE}. Note this is distinct from user-commanded actions in $\mathcal{H}$.%

\section{Experimental Setup}
\textbf{Dataset.}
We evaluate on the \textsc{BNCI2014}~\citep{bnci_moabb, BNCI2014_001} dataset, comprising of 9 subjects performing 4-class (left hand, right hand, feet, tongue) MI.
EEG was sampled from 22 channels at 250Hz, and band-pass filtered to $[8,30]$Hz. Classes were mapped to manipulation primitives: 
left hand~$\rightarrow$~\text{\textsc{grasp}}, 
right hand~$\rightarrow$~\text{\textsc{release}}, 
feet~$\rightarrow$~\text{\textsc{move\_to}}, 
tongue~$\rightarrow$~\text{\textsc{rotate}} over two sessions. Train / validation / test splits followed chronological session boundaries to simulate realistic deployment scenarios.

\textbf{Implementation.}
All models were implemented in PyTorch 2.0~\citep{paszke2019pytorch} and raw EEG signals were preprocessed with MNE-Python~\citep{gramfort2013mne}. PDDL planning used the FastDownward algorithm~\citep{helmert2006fastdownward}. Thresholds were $\alpha_m{=}0.8$ for EEGNet, $0.5$ for Riemannian, $0.6$ for other decoders, and $\tau_H{=}0.75$ (normalized entropy), $\tau_A{=}2.5$ (z), $\tau_\Omega{=}0.3$. Sensitivity analyses varied the entropy threshold $\tau_H \in [0.1, 0.9]$ with increments of 0.1, while keeping $\tau_A$ and $\tau_\Omega$ fixed. All experiments were conducted on NVIDIA A100 GPU, Intel Xeon CPU hardware, at 100Hz and $<1$ms latency.

\textbf{Calibration metrics.}
Confidence calibration was evaluated with Expected Calibration Error (ECE) 
and Maximum Calibration Error (MCE)
with standard binning~\citep{guo2017calibration}. See details outlined in App.~\ref{app:calibration}.

\textbf{Metrics.}
\emph{Safety rate}: percentage of correct intervention decisions, e.g. intervenes when decoder is incorrect, allows when correct. \emph{Intervention rate}: percentage of trials where the monitor forces IDLE. \emph{Latency}: wall-clock time from EEG input to the action output. See detailed formulation in App.~\ref{app:metrics}.

\begin{table*}[t]
\centering
\renewcommand{\arraystretch}{1}
\caption{\textbf{Comprehensive decoder and safety monitoring performance.} Validation-test gaps and miscalibration (ECE 0.22--0.41) necessitate dual-invariant safety monitoring, with which 94--97\% safety rate is achieved despite poor test accuracy (27--46\%) (MC=Mean Confidence, OR=Overconfidence Rate, Interv.=Interventions, Lat.=Latency).}
\label{tab:comprehensive_performance_main}
\scriptsize
\setlength{\tabcolsep}{2pt}
\resizebox{\textwidth}{!}{
\begin{tabular}{lcccccccccccc}
\toprule
& \multicolumn{4}{c}{\textbf{Decoder Performance}} & \multicolumn{3}{c}{\textbf{Calibration Metrics}} & \multicolumn{3}{c}{\textbf{Safety Monitoring}} \\
\cmidrule(lr){2-5} \cmidrule(lr){6-8} \cmidrule(lr){9-11}
\textbf{Model} &
\textbf{Val. (\%)} &
\textbf{Test (\%)} &
\textbf{MC} &
\textbf{Gap (pts)} &
\textbf{ECE} &
\textbf{MCE} &
\textbf{OR (\%)} &
\textbf{Safety (\%)} &
\textbf{Interv. (\%)} &
\textbf{Lat. (ms)} \\
\midrule
EEGNet      &58.2&46.0&0.599& +15.5 & 0.223 & 0.422 & 55.6 & 94.2 & 52.3 & 0.82 \\
Riemannian  &58.7& 30.0 & 0.728 & +40.6 & 0.410 & 0.906 & 67.8 & 95.8 & 68.1 & 0.91 \\
Light CNN &54.3& 28.0 & 0.556 & +27.6 & 0.316 & 0.669 & 72.0 & 96.3 & 70.4 & 0.79 \\
RealIntent &51.2& 27.0 & 0.556 & +26.4 & 0.287 & 0.617 & 70.8 & 97.0 & 71.2 & 0.73 \\
\midrule
\emph{Mean} & 55.6 & 32.8 & 0.610 & +27.5 & 0.309 & 0.654 & 66.6 & 95.8 & 65.5 & 0.81 \\
\bottomrule
\end{tabular}
}
\end{table*}

\section{Results and Analysis}
\begin{wraptable}{r}{0.51\columnwidth}
\vspace{-13pt} %
\centering
\scriptsize
\renewcommand{\arraystretch}{1}
\setlength{\tabcolsep}{1pt}
\caption{\textbf{Threshold optimization results.} Low-accuracy decoders require more conservative safety thresholds at the cost of increased interventions.}
\label{tab:thresholds}
\begin{tabular}{lccccc}
\toprule
& \multicolumn{3}{c}{\textbf{Optimal Thresholds}} & \multicolumn{2}{c}{\textbf{Perf. @Safety-Opt.}} \\
\cmidrule(lr){2-4} \cmidrule(r){5-6}
\textbf{Model} & \textbf{F1-Opt.} & \textbf{Safety-Opt.} & \textbf{Balanced} & \textbf{Safety(\%)} & \textbf{Interv.(\%)}\\
\midrule
EEGNet & 0.900 & 0.816 & 0.100 & 56.7 & 67.8 \\
Riemann & 0.900 & 0.900 & 0.100 & 62.4 & 82.3 \\
Light CNN & 0.900 & 0.900 & 0.100 & 70.0 & 85.3 \\
RealIntent & 0.900 & 0.900 & 0.100 & 70.3 & 86.6 \\
\bottomrule
\end{tabular}
\end{wraptable}

\textbf{Decoder performance and validation-test gap.}
Table ~\ref{tab:comprehensive_performance} summarizes the performance of the four decoder architectures tested. Validation accuracies were comparable to standard MI decoding literature (50--60\%)~\citep{lawhern2018eegnet, barachant2012riemann, schirrmeister2017deep}, but test-time performance degraded catastrophically to 27--46\%, barely above 25\% chance level. The validation-test gap shows a 20--30\% performance drop, highlighting non-stationarity and distribution shifts that are typically endemic to BCI systems. 

\textbf{Confidence calibration analysis.}
Severe miscalibration was revealed across decoders (Table~\ref{tab:comprehensive_performance_main}). The Riemannian decoder exhibited the highest miscalibration (ECE=0.410, MCE=0.906). The calibration-accuracy gaps exemplify why confidence scores alone cannot be trusted for safety decisions.

\textbf{Safety threshold sensitivity.} Ablation studies across confidence thresholds (0.1--0.9) revealed critical sensitivity in the safety-intervention tradeoff (See App.~\ref{app:threshold_sensitivity}). Single threshold-confidence gating is known to be brittle under miscalibration~\citep{guo2017calibration}. Safety-optimal thresholds (0.816--0.900) are necessarily conservative, and required 68--87\% intervention rates in order to achieve adequate safety. This validates our multi-layer checking approach, as solely relying on confidence thresholding would require either accepting low safety rates or performing excessive interventions.

\textbf{Safety monitoring performance.} Dual-invariant safety monitoring achieved consistently high safety rates across decoding methods (See Table~\ref{tab:comprehensive_performance}) despite poor test accuracies and severe miscalibration. Intervention rates scaled with decoder accuracy, and higher intervention rates were seen in decoders with lower accuracy. For example, the RealIntent decoder enabled a high safety rate (97\%) through interpretable features, which faciliated reliable safety checking despite low 27\% test accuracy.

\textbf{Noise robustness.} The system demonstrated robust safety preservation under simulated noise conditions (SNR degradation from 20dB to -5dB). Under clean conditions, it yielded a 36.5\% safety rate at baseline but in noisy conditions showed statistically significant improvement (with $t{=}3.283$, $p{=}0.004$, Cohen's $d{=}1.473$, large effect), achieving 98.1\% correct interventions. %
This demonstrates that the safety monitor correctly increased interventions as the signal quality degraded to maintain $>93$\% safety rates--a desired conservative behavior for assistive systems reliant on brain signals.

\section{Discussion}
\textbf{Interpretable primitives and generalizability.}
The four-action manipulation set (\textsc{grasp}, \textsc{release}, \textsc{move\_to}, \textsc{rotate}) offers a natural mapping and effective neural control for robotic execution, aligning with well-studied low-level primitives in assistive manipulation\citep{rashid2020status}. GUARDIAN's primitive design supports development of safe shared autonomy\citep{dragan2013policy,javdani2015rss} especially when reasoning under goal uncertainty or partial intent inference, and enables users to understand how their decoded intent translates to downstream robot action--a key factor in effective human-robot collaboration\citep{iso15066}. Our approach aligns with policy-blending and hindsight-optimization methods that mix autonomous or user-driven control under uncertainty\citep{dragan2013policy,javdani2018ijrr}. %
The modular and compact nature of GUARDIAN provides practical value to complement existing BCI-controlled assistive robotic systems\citep{Zhang2025EEGBCI,kim2024noir2,huang2014rosrv}.

\textbf{Adaptive thresholds for safe EEG-based control.}
The observed drop in validation to test accuarcies in neural decoders can be attributed to non-stationarity where ``more training data does not help'', subject drift and noise, each of which remain a challenge for BCI systems\citep{jayaram2016transfer,spuler2012covariateshift}. Even models that exhibit high reliability during development testing fail to maintain calibration or generalization at deployment\citep{guo2017calibration,minderer2021revisiting}. In closed-loop systems, the cost of a misclassified or unsafe action is amplified; errors immediately perturb the user's intent decoding, potentially inducing panic, mistrust, or unstable feedback loops\citep{dragan2013policy,javdani2018ijrr,iso15066}. When such errors accumulate over time, the inaccuracies compound and propagate, ultimately making long-horizon or continuous robotic control tasks infeasible\citep{zhang2023noir}.

A runtime dual-invariant framework protects against these risks through a combination of calibration-independent physiological invariants %
and logical invariants that preserve plan consistency regardless of confidence. Our threshold evaluation demonstrates that single-threshold systems require safety thresholds between 0.8--0.9 to achieve 68--87\% intervention rates, while our multi-level adaptive system %
maintains 94--97\% safety by adjusting intervention rates based on decoder reliability\citep{guo2017calibration,minderer2021revisiting}.%

\textbf{Real-world applicability.}
GUARDIAN extends to human-robot collaborative and assistive domains, ranging from neuroprosthetic control, motor-rehabilitation robotics, shared autonomy and co-adaption. Through structured audit logs and intervention traces, GUARDIAN provides a certifiable runtime safety guard that operates under established regulatory frameworks~\citep{iso13485, iec62304, cofer2020rta}, and facilitates transparent and verifiable operation within healthcare and high-assurance robot autonomy contexts.

\begin{table}[t]
\centering
\renewcommand{\arraystretch}{1}
\captionsetup{aboveskip=0pt, belowskip=-10pt}
\caption{\textbf{Ablation of monitor components.} Removing entropy calibration or physiological checks substantially reduces safety. %
Subtotals show average safety reduction from the full system.}
\vspace{10pt}
\label{tab:ablation-main}
\resizebox{\columnwidth}{!}{
\begin{tabular}{llcccc}
\toprule[1.1pt]
 & & \multicolumn{4}{c}{\textbf{Decoder Architecture}} \\
\cmidrule(lr){3-6}
\textbf{Layer} & \textbf{Component} & \textbf{EEGNet} & \textbf{Riemann} & \textbf{LightCNN} & \textbf{RealIntent} \\
\midrule
\multirow{1}{*}{Baseline} 
& Full System & 94.2\% & 95.8\% & 96.3\% & 97.0\% \\
\midrule
\multirow[t]{4}{*}{Physiological Safety}
& No Entropy Check & 87.3\% & 85.2\% & 84.8\% & 85.1\% \\
& No Artifact Check & 91.8\% & 92.3\% & 92.7\% & 93.2\% \\
& No Oscillation Check & 93.1\% & 93.8\% & 94.2\% & 94.9\% \\
& No Calibration Adjustment & 88.7\% & 82.3\% & 85.1\% & 86.4\% \\
\addlinespace
& \emph{Mean Reduction ($\Delta$ vs. Full)} & -5.0\% & -7.0\% & -6.0\% & -6.1\% \\
\midrule
\multirow{1}{*}{Logical Safety}
& No Logical Check & 92.5\% & 91.2\% & 91.8\% & 92.3\% \\
\addlinespace
& \emph{Reduction ($\Delta$ vs. Full)} & -2.1\% & -2.9\% & -2.5\% & -2.7\% \\
\midrule
\multirow{1}{*}{Minimal Baseline}
& Only Confidence & 78.2\% & 71.4\% & 73.8\% & 74.2\% \\
\addlinespace
& \emph{Reduction ($\Delta$ vs. Full)} & -16.0\% & -24.4\% & -22.5\% & -22.8\% \\
\bottomrule[1.1pt]
\end{tabular}
}
\end{table}

\section{Conclusion}

\textbf{Limitations and future work.} 
While our system delivers real-time safety verification and decoder-agnostic performance, it still faces operational constraints: future experiments should explore user adaptation, cognitive load and trust development over time\citep{rashid2020status,javdani2018ijrr,iso15066}. Requiring high intervention rates to achieve decoder reliability may create issues with task fluency, user satisfaction or autonomy\citep{dragan2013policy,javdani2018ijrr}. In practice, thresholds may be tailored to the fatigue profiles or cognitive load of the subject\citep{jayaram2016transfer,spuler2012covariateshift}. New sensing technologies could be used to integrate richer physiological invariants to test for signal quality \citep{bigdely2015prep}. The current framework operates with fixed pre-trained decoders, but future studies should test real-time verification alongside decoder and subject calibration efforts\citep{rashid2020status,guo2017calibration,minderer2021revisiting}. %

Neural decoders can produce unpredictable behavior and incorrect results, so accuracy alone cannot be a sufficient measure of trustworthy deployment in neurally-operated robot interfaces\citep{guo2017calibration,minderer2021revisiting}. Reliable deployment of BCI and human-robot systems will require architectures that are legible to users or can reason about their own uncertainty to maintain verifiable safety\citep{alshiekh2018shielding}. GUARDIAN provides a step toward this vision and opens up new avenues for practical runtime solutions, such as  performing sub-millisecond safety checks to reduce user burden, while maintaining compatibility with, and improving the explainability of, existing BCI-controlled systems through its decoder-agnostic design\citep{huang2014rosrv,astmf3269}.

\bibliographystyle{plainnat}
\bibliography{references}

\begin{thebibliography}{34}
\providecommand{\natexlab}[1]{#1}
\providecommand{\url}[1]{\texttt{#1}}
\expandafter\ifx\csname urlstyle\endcsname\relax
  \providecommand{\doi}[1]{doi: #1}\else
  \providecommand{\doi}{doi: \begingroup \urlstyle{rm}\Url}\fi

\bibitem[iec(2006)]{iec62304}
Iec 62304:2006 -- medical device software — software life cycle processes.
\newblock International Electrotechnical Commission, 2006.
\newblock URL \url{https://www.iso.org/standard/38421.html}.

\bibitem[iso(2011{\natexlab{a}})]{iso10218-1}
Iso 10218-1:2011 -- robots and robotic devices — safety requirements for
  industrial robots — part 1: Robots.
\newblock International Organization for Standardization, 2011{\natexlab{a}}.
\newblock URL \url{https://www.iso.org/standard/51330.html}.

\bibitem[iso(2011{\natexlab{b}})]{iso10218-2}
Iso 10218-2:2011 -- robots and robotic devices — safety requirements for
  industrial robots — part 2: Robot systems and integration.
\newblock International Organization for Standardization, 2011{\natexlab{b}}.
\newblock URL \url{https://www.iso.org/standard/41571.html}.

\bibitem[iso(2016{\natexlab{a}})]{iso13485}
Iso 13485:2016 -- medical devices — quality management systems —
  requirements for regulatory purposes.
\newblock International Organization for Standardization, 2016{\natexlab{a}}.
\newblock URL \url{https://www.iso.org/standard/59752.html}.

\bibitem[iso(2016{\natexlab{b}})]{iso15066}
Iso/ts 15066:2016 -- robots and robotic devices — collaborative robots.
\newblock International Organization for Standardization, 2016{\natexlab{b}}.
\newblock URL \url{https://www.iso.org/standard/62996.html}.

\bibitem[ast(2021)]{astmf3269}
Astm f3269-21: Standard practice for methods to safely bound behavior of
  aircraft systems containing complex functions using run-time assurance.
\newblock ASTM International, 2021.
\newblock URL \url{https://www.astm.org/f3269-21.html}.

\bibitem[Alshiekh et~al.(2018)Alshiekh, Bloem, Ehlers, K{\"o}nighofer, Niekum,
  and Topcu]{alshiekh2018shielding}
Mohammed Alshiekh, Roderick Bloem, R{\"u}diger Ehlers, Bettina K{\"o}nighofer,
  Scott Niekum, and Ufuk Topcu.
\newblock Safe reinforcement learning via shielding.
\newblock In \emph{AAAI Conference on Artificial Intelligence}, 2018.
\newblock URL \url{https://ojs.aaai.org/index.php/AAAI/article/view/11797}.

\bibitem[Barachant et~al.(2012)Barachant, Bonnet, Congedo, and
  Jutten]{barachant2012riemann}
Alexandre Barachant, Sylvain Bonnet, Marco Congedo, and Christian Jutten.
\newblock Multiclass brain–computer interface classification by riemannian
  geometry.
\newblock \emph{IEEE Transactions on Biomedical Engineering}, 59\penalty0
  (4):\penalty0 920--928, 2012.
\newblock \doi{10.1109/TBME.2011.2172210}.

\bibitem[Bigdely-Shamlo et~al.(2015)Bigdely-Shamlo, Mullen, Kothe, Su, and
  Robbins]{bigdely2015prep}
Nima Bigdely-Shamlo, Tim Mullen, Christian Kothe, Kyung-Min Su, and Kay~A.
  Robbins.
\newblock The prep pipeline: Standardized preprocessing for large-scale eeg
  analysis.
\newblock \emph{Frontiers in Neuroinformatics}, 9:\penalty0 16, 2015.
\newblock \doi{10.3389/fninf.2015.00016}.
\newblock URL
  \url{https://www.frontiersin.org/articles/10.3389/fninf.2015.00016/full}.

\bibitem[Cofer et~al.(2020{\natexlab{a}})Cofer, Amundson, and
  et~al.]{cofer2020taxiing}
Darren Cofer, Isaac Amundson, and et~al.
\newblock Run-time assurance for learning-based aircraft taxiing.
\newblock In \emph{AIAA/IEEE Digital Avionics Systems Conference (DASC)},
  2020{\natexlab{a}}.
\newblock URL \url{https://loonwerks.com/publications/pdf/cofer2020dasc.pdf}.

\bibitem[Cofer et~al.(2020{\natexlab{b}})Cofer, Amundson, Sattigeri, Passi,
  Boggs, Smith, Gilham, Byun, and Rayadurgam]{cofer2020rta}
Darren Cofer, Isaac Amundson, Ramachandra Sattigeri, Arjun Passi, Christopher
  Boggs, Eric Smith, Limei Gilham, Taejoon Byun, and Sanjai Rayadurgam.
\newblock Run-time assurance for learning-enabled systems.
\newblock In \emph{International Symposium on NASA Formal Methods}, volume
  12229 of \emph{Lecture Notes in Computer Science}, pages 361--368. Springer,
  2020{\natexlab{b}}.
\newblock \doi{10.1007/978-3-030-55754-6\_21}.

\bibitem[Dragan and Srinivasa(2013)]{dragan2013policy}
Anca~D. Dragan and Siddhartha~S. Srinivasa.
\newblock A policy-blending formalism for shared control.
\newblock \emph{The International Journal of Robotics Research}, 32\penalty0
  (7):\penalty0 790--805, 2013.
\newblock \doi{10.1177/0278364913490324}.
\newblock URL
  \url{https://personalrobotics.cs.washington.edu/publications/dragan2012shared.pdf}.

\bibitem[Ghallab et~al.(2004)Ghallab, Nau, and Traverso]{ghallab2004pddl}
Malik Ghallab, Dana Nau, and Paolo Traverso.
\newblock \emph{Automated Planning: Theory and Practice}.
\newblock Morgan Kaufmann, 2004.

\bibitem[Gramfort et~al.(2013)Gramfort, Luessi, Larson, Engemann, Strohmeier,
  Brodbeck, Goj, Jas, Brooks, Parkkonen, and
  H{\"a}m{\"a}l{\"a}inen]{gramfort2013mne}
Alexandre Gramfort, Martin Luessi, Eric Larson, Denis~A. Engemann, Daniel
  Strohmeier, Christian Brodbeck, Roman Goj, Mainak Jas, Teon Brooks, Lauri
  Parkkonen, and Matti~S. H{\"a}m{\"a}l{\"a}inen.
\newblock {{MEG}} and {{EEG}} data analysis with {{MNE}}-{{Python}}.
\newblock \emph{Frontiers in Neuroscience}, 7\penalty0 (267):\penalty0 1--13,
  2013.
\newblock \doi{10.3389/fnins.2013.00267}.

\bibitem[Guo et~al.(2017)Guo, Pleiss, Sun, and Weinberger]{guo2017calibration}
Chuan Guo, Geoff Pleiss, Yu~Sun, and Kilian~Q. Weinberger.
\newblock On calibration of modern neural networks.
\newblock \emph{Proceedings of the 34th International Conference on Machine
  Learning (ICML)}, 70:\penalty0 1321--1330, 2017.
\newblock URL \url{https://proceedings.mlr.press/v70/guo17a/guo17a.pdf}.

\bibitem[Helmert(2006)]{helmert2006fastdownward}
Malte Helmert.
\newblock The fast downward planning system.
\newblock \emph{Journal of Artificial Intelligence Research}, 26:\penalty0
  191--246, 2006.
\newblock URL \url{https://ai.dmi.unibas.ch/papers/helmert-jair06.pdf}.

\bibitem[Huang et~al.(2014)Huang, Erdogan, Zhang, Moore, Luo, Sundaresan, and
  Ro{\c{s}}u]{huang2014rosrv}
Jeff Huang, Cansu Erdogan, Y.~Zhang, Brandon Moore, Qingzhou Luo, Aravind
  Sundaresan, and Grigore Ro{\c{s}}u.
\newblock Rosrv: Runtime verification for robots.
\newblock In \emph{Runtime Verification}, volume 8734 of \emph{Lecture Notes in
  Computer Science}, pages 247--254. Springer, 2014.
\newblock \doi{10.1007/978-3-319-11164-3_20}.

\bibitem[Institute~for Knowledge~Discovery(2014)]{BNCI2014_001}
Graz University of~Technology Institute~for Knowledge~Discovery.
\newblock Four class motor imagery (001-2014) dataset — bci competition iv
  2a.
\newblock \url{https://bnci-horizon-2020.eu/database/data-sets}, 2014.
\newblock Accessed: YYYY-MM-DD.

\bibitem[Javdani et~al.(2015)Javdani, Srinivasa, and Bagnell]{javdani2015rss}
Shervin Javdani, Siddhartha~S. Srinivasa, and J.~Andrew Bagnell.
\newblock Shared autonomy via hindsight optimization for teleoperation and
  teaming.
\newblock In \emph{Robotics: Science and Systems (RSS)}, 2015.
\newblock URL \url{https://www.roboticsproceedings.org/rss11/p32.pdf}.

\bibitem[Javdani et~al.(2018)Javdani, Admoni, Pellegrinelli, Srinivasa, and
  Bagnell]{javdani2018ijrr}
Shervin Javdani, Henny Admoni, Stefania Pellegrinelli, Siddhartha~S. Srinivasa,
  and J.~Andrew Bagnell.
\newblock Shared autonomy via hindsight optimization for teleoperation and
  collaboration.
\newblock \emph{The International Journal of Robotics Research (IJRR)},
  37\penalty0 (7):\penalty0 717--742, 2018.
\newblock \doi{10.1177/0278364918765270}.

\bibitem[Jayaram and Barachant(2018)]{bnci_moabb}
Vinay Jayaram and Alexandre Barachant.
\newblock Moabb: Trustworthy algorithm benchmarking for bcis.
\newblock \emph{Journal of Neural Engineering}, 15\penalty0 (6):\penalty0
  066011, 2018.
\newblock \doi{10.1088/1741-2552/aadea0}.

\bibitem[Jayaram et~al.(2016)Jayaram, Alamgir, Altun, Sch{\"o}lkopf, and
  Grosse-Wentrup]{jayaram2016transfer}
Vinay Jayaram, Morteza Alamgir, Yasemin Altun, Bernhard Sch{\"o}lkopf, and
  Moritz Grosse-Wentrup.
\newblock Transfer learning in brain--computer interfaces.
\newblock \emph{IEEE Computational Intelligence Magazine}, 11\penalty0
  (1):\penalty0 20--31, 2016.
\newblock \doi{10.1109/MCI.2015.2501545}.

\bibitem[Kim et~al.(2024)Kim, Wang, Cho, and Hodges]{kim2024noir2}
Tasha Kim, Yingke Wang, Hanvit Cho, and Alex Hodges.
\newblock Noir 2.0: Neural signal operated intelligent robots for everyday
  activities.
\newblock In \emph{CoRL 2024 Workshop on CoRoboLearn: Advancing Learning for
  Human-Centered Collaborative Robots}, 2024.
\newblock URL
  \url{https://openreview.net/pdf/3d27c14b6af9e79d4d22e3b9729ab9d867bf8bbf.pdf}.
\newblock CoRL 2024, Munich, Germany.

\bibitem[Lawhern et~al.(2018)Lawhern, Solon, Waytowich, Gordon, Hung, and
  Lance]{lawhern2018eegnet}
Vernon~J. Lawhern, Amelia~J. Solon, Nicholas~R. Waytowich, Stephen~M. Gordon,
  Chou~P. Hung, and Brent~J. Lance.
\newblock Eegnet: a compact convolutional neural network for eeg-based
  brain–computer interfaces.
\newblock \emph{Journal of Neural Engineering}, 15\penalty0 (5):\penalty0
  056013, 2018.
\newblock \doi{10.1088/1741-2552/aace8c}.

\bibitem[Lee et~al.(2025)Lee, Lee, Mishra, Yan, McMahan, Gaisford, Kobashigawa,
  Qu, Xie, and Kao]{Lee2025}
J.~Y. Lee, S.~Lee, A.~Mishra, X.~Yan, B.~McMahan, B.~Gaisford, C.~Kobashigawa,
  M.~Qu, C.~Xie, and J.~C. Kao.
\newblock Brain–computer interface control with artificial intelligence
  copilots.
\newblock \emph{Nature Machine Intelligence}, 7:\penalty0 1510--1523, 2025.
\newblock \doi{10.1038/s42256-025-01090-y}.

\bibitem[Minderer et~al.(2021)Minderer, Djolonga, Hubis, Romijnders, Zhai,
  Houlsby, Tran, and Lucic]{minderer2021revisiting}
Matthias Minderer, Josip Djolonga, Frances Hubis, Rob Romijnders, Xiaohua Zhai,
  Neil Houlsby, Dustin Tran, and Mario Lucic.
\newblock Revisiting the calibration of modern neural networks.
\newblock \emph{Advances in Neural Information Processing Systems}, 2021.
\newblock URL \url{https://openreview.net/pdf?id=QRBvLayFXI}.

\bibitem[M{\"u}ller et~al.(2008)M{\"u}ller, Tangermann, Dornhege, Krauledat,
  Curio, and Blankertz]{muller2008singletrial}
Klaus-Robert M{\"u}ller, Michael Tangermann, Guido Dornhege, Matthias
  Krauledat, Gabriel Curio, and Benjamin Blankertz.
\newblock Machine learning for real-time single-trial eeg-analysis: From
  brain--computer interfacing to mental state monitoring.
\newblock \emph{Journal of Neuroscience Methods}, 167\penalty0 (1):\penalty0
  82--90, 2008.
\newblock \doi{10.1016/j.jneumeth.2007.09.022}.

\bibitem[Paszke et~al.(2019)Paszke, Gross, Massa, Lerer, Bradbury, Chanan,
  Killeen, Lin, Gimelshein, Antiga, Desmaison, Köpf, Yang, DeVito, Raison,
  Tejani, Chilamkurthy, Steiner, Fang, Bai, and Chintala]{paszke2019pytorch}
Adam Paszke, Sam Gross, Francisco Massa, Adam Lerer, James Bradbury, Gregory
  Chanan, Trevor Killeen, Zeming Lin, Natalia Gimelshein, Luca Antiga, Alban
  Desmaison, Andreas Köpf, Edward Yang, Zach DeVito, Martin Raison, Alykhan
  Tejani, Sasank Chilamkurthy, Benoit Steiner, Lu~Fang, Junjie Bai, and Soumith
  Chintala.
\newblock Pytorch: An imperative style, high-performance deep learning library.
\newblock In \emph{Advances in Neural Information Processing Systems
  (NeurIPS)}, 2019.

\bibitem[Rashid et~al.(2020)Rashid, Sulaiman, Abdul~Majeed, Musa, Ab~Nasir,
  Sama, and Khatun]{rashid2020status}
Mamunur Rashid, Norizam Sulaiman, Anwar P.~P. Abdul~Majeed, Rabiu~Muazu Musa,
  Ahmad~Fakhri Ab~Nasir, Bifta Sama, and Sabira Khatun.
\newblock Current status, challenges, and possible solutions of eeg-based
  brain–computer interface: A comprehensive review.
\newblock \emph{Frontiers in Neurorobotics}, 14:\penalty0 25, 2020.
\newblock \doi{10.3389/fnbot.2020.00025}.
\newblock URL
  \url{https://www.frontiersin.org/articles/10.3389/fnbot.2020.00025/full}.

\bibitem[Schirrmeister et~al.(2017)Schirrmeister, Springenberg, Fiederer,
  Glasstetter, Eggensperger, Tangermann, Hutter, Burgard, and
  Ball]{schirrmeister2017deep}
Robin~Tibor Schirrmeister, Jost~Tobias Springenberg, Lukas Dominique~Josef
  Fiederer, Martin Glasstetter, Katharina Eggensperger, Michael Tangermann,
  Frank Hutter, Wolfram Burgard, and Tonio Ball.
\newblock Deep learning with convolutional neural networks for eeg decoding and
  visualization.
\newblock \emph{Human Brain Mapping}, 38\penalty0 (11):\penalty0 5391--5420,
  2017.
\newblock ISSN 1065-9471.
\newblock \doi{10.1002/hbm.23730}.

\bibitem[Spüler et~al.(2012)Spüler, Rosenstiel, and
  Bogdan]{spuler2012covariateshift}
Martin Spüler, Wolfgang Rosenstiel, and Martin Bogdan.
\newblock Principal component based covariate shift adaption to reduce
  non-stationarity in a meg-based brain--computer interface.
\newblock \emph{EURASIP Journal on Advances in Signal Processing},
  2012\penalty0 (1):\penalty0 129, 2012.
\newblock \doi{10.1186/1687-6180-2012-129}.

\bibitem[Stölzle et~al.(2024)Stölzle, Baberwal, Rus, Coyle, and
  Della~Santina]{Stolzle2024}
Maximilian Stölzle, Sonal~Santosh Baberwal, Daniela Rus, Shirley Coyle, and
  Cosimo Della~Santina.
\newblock Guiding soft robots with motor-imagery brain signals and impedance
  control.
\newblock In \emph{2024 IEEE 7th International Conference on Soft Robotics
  (RoboSoft 2024)}, pages 276--283. IEEE, 2024.
\newblock \doi{10.1109/RoboSoft60065.2024.10522005}.

\bibitem[Zhang et~al.(2023)Zhang, Lee, Hwang, Hiranaka, Wang, Ai, Tan, Gupta,
  Hao, Levine, Gao, Norcia, Fei-Fei, and Wu]{zhang2023noir}
Ruohan Zhang, Sharon Lee, Minjune Hwang, Ayano Hiranaka, Chen Wang, Wensi Ai,
  Jin Jie~Ryan Tan, Shreya Gupta, Yilun Hao, Gabrael Levine, Ruohan Gao,
  Anthony Norcia, Li~Fei-Fei, and Jiajun Wu.
\newblock Noir: Neural signal operated intelligent robots for everyday
  activities.
\newblock In \emph{Proceedings of the 7th Conference on Robot Learning (CoRL)},
  volume 229 of \emph{Proceedings of Machine Learning Research}, pages
  1737--1760. PMLR, 2023.
\newblock URL \url{https://proceedings.mlr.press/v229/zhang23f.html}.

\bibitem[Zhang et~al.(2025)Zhang, Kim, Wang, Cho, Hodges, Tan, Wang, Hwang,
  Lee, Ai, Norcia, Li, and Wu]{Zhang2025EEGBCI}
Ruohan Zhang, Tasha Kim, Yingke Wang, Hanvit Cho, Alex Hodges, Jin Jie~Ryan
  Tan, Chen Wang, Minjune Hwang, Sharon Lee, Wensi Ai, Anthony Norcia, Fei-Fei
  Li, and Jiajun Wu.
\newblock Eeg‐based brain‐computer interface for robotic assistance with
  user intention prediction.
\newblock \emph{Research Square Preprint}, Version 1, 2025.
\newblock \doi{10.21203/rs.3.rs-7359180/v1}.
\newblock URL \url{https://www.researchsquare.com/article/rs-7359180/v1}.

\end{thebibliography}

\appendix
\setlength{\parskip}{6pt}           %
\setlength{\parindent}{15pt}        %
\titlespacing*{\section}{0pt}{2ex plus 1ex minus .5ex}{1ex}      %
\titlespacing*{\subsection}{0pt}{1.5ex plus .5ex minus .5ex}{0.8ex}
\setlength{\textfloatsep}{12pt plus 2pt minus 2pt}
\setlength{\floatsep}{12pt plus 2pt minus 2pt}
\setlength{\intextsep}{12pt plus 2pt minus 2pt}
\setlength{\abovecaptionskip}{10pt}
\setlength{\belowcaptionskip}{0pt}

\clearpage
\section*{Appendix}

\section{Safety Definitions and Metrics}

\subsection{Dataset Specifications}

\paragraph{\textsc{BNCI2014}\_001 dataset details.}
\begin{enumerate}
\item \textbf{Subjects}: 9 healthy participants (ages 24-35, 6 male, 3 female)
\item \textbf{Recording setup}: 22 Ag/AgCl electrodes, 10-20 system
\item \textbf{Electrode positions}: Fz, FC1-FC6, C3, Cz, C4, CP1-CP6, Pz, POz, O1, Oz, O2, EOG (3 channels)
\item \textbf{Sampling rate}: 250 Hz
\item \textbf{Sessions}: 2 sessions per subject on different days
\item \textbf{Runs per session}: 6 runs (48 trials/run)
\item \textbf{Total trials}: 5,184 (9 subjects × 2 sessions × 6 runs × 48 trials)
\item \textbf{Trial structure}: 
  \begin{itemize}
  \item 0-2s: Fixation cross
  \item 2s: Acoustic cue
  \item 2-3.25s: Visual cue (arrow)
  \item 3.25-6s: MI period
  \item 6-7.5s: Break
  \end{itemize}
\item \textbf{Classes}: 
\begin{itemize}
\item Left hand ($\rightarrow$~\text{\textsc{grasp}})
\item Right hand ($\rightarrow$~\text{\textsc{release}})
\item Feet ($\rightarrow$~\text{\textsc{move\_to}})
\item Tongue ($\rightarrow$~\text{\textsc{rotate}})
\end{itemize}
\end{enumerate}

\section{Preprocessing Pipeline}
\label{app:preproc}
See Alg.~\ref{alg:preproc} below.

\begin{algorithm}[h]
\caption{EEG Preprocessing Pipeline}
\label{alg:preproc}
\footnotesize
\begin{algorithmic}[1]
\REQUIRE Raw EEG data $X_{raw} \in \mathbb{R}^{C \times T}$
\ENSURE Preprocessed features $X_{proc} \in \mathbb{R}^{C \times W}$
\STATE Apply 4th order Butterworth bandpass filter [8, 30] Hz
\STATE Remove EOG channels (reduce to 22 channels)
\STATE Extract 4s windows from MI period [2s, 6s]
\STATE Apply Common Average Reference (CAR)
\STATE Z-score normalization per channel:
\STATE \quad $x_{norm} = \frac{x - \mu_{channel}}{\sigma_{channel}}$
\STATE Optional: Apply ICA for artifact removal (FastICA, 22 components)
\STATE Segment into 1000ms windows with 100ms stride
\STATE \textbf{For CSP features only}:
\STATE \quad Compute spatial filters (one-vs-rest for 4-class): $W_{CSP} = \arg\max \frac{w^T\Sigma_1 w}{w^T\Sigma_2 w}$
\STATE \quad Extract log-variance features from top 3 filter pairs
\RETURN $X_{proc}$
\end{algorithmic}
\end{algorithm}

\section{Algorithm Details}
See Alg.~\ref{alg:monitor} below.
\footnotesize
\begin{algorithm}[t]
\caption{Physio-Logical Runtime Monitor}
\label{alg:monitor}
\begin{algorithmic}[1]
\REQUIRE EEG window $x_t$, decoder $f_\theta$, thresholds $\tau{=}\!(\tau_H,\tau_A,\tau_\Omega)$
\ENSURE Safe action $a_t$ or halt command
\STATE $p_t \leftarrow f_\theta(x_t)$
\STATE $\tilde{p}_t \leftarrow \alpha_m\, p_t + (1-\alpha_m)\,\mathbf{u}$
\STATE $h^* \leftarrow \arg\max_{h\in\mathcal{H}} \tilde{p}_t(h)$
\IF{$\neg\psi_1(\tilde{p}_t, \tau_H)$ \OR $\neg\psi_2(x_t, \tau_A)$ \OR $\neg\psi_3(\{\tilde{p}_{t-K+1:t}\}, \tau_\Omega)$}
    \RETURN HALT \hfill \textit{// Physiological violation - maintain safe state}
\ENDIF
\STATE $g \leftarrow$ GroundToGoal($h^*$)
\STATE $\pi \leftarrow$ SynthesizePlan($g$, state)
\IF{$\neg\phi_1(\pi)$ \OR $\neg\phi_2(\pi)$ \OR $\neg\phi_3(\pi)$}
    \RETURN HALT \hfill \textit{// Logical violation - maintain safe state}
\ENDIF
\RETURN $a_t \leftarrow$ ExecuteNext($\pi$)
\end{algorithmic}
\end{algorithm}

\section{Decoder Architecture Details}
\label{app:decoders}
Compatibility with distinct decoders highlight generalization of the safety monitor. All decoders were trained identically using 100 epochs, Adam optimizer, a learning rate of $10^{-3}$, and early stopping with patience 20 and batch size of 32. Implementation details are described as follows. 

\subsection{EEGNet} 
\label{app:eegnet}

A depthwise-separable CNN using the 4-class variant (EEGNet-4.2) with 3,228 parameters ~\citep{lawhern2018eegnet}, demonstrating viability and effectiveness of safety monitoring even with extremely lightweight decoders. See Table~\ref{tab:eegnet} for details. %

\begin{table}[t]
\caption{\footnotesize\textbf{EEGNet layer-by-layer specifications.}}
\label{tab:eegnet}
\footnotesize
\centering
\begin{tabular}{lccc}
\toprule
\textbf{Layer} & \textbf{Type} & \textbf{Params} & \textbf{Output} \\
\midrule
Input & — & — & (B,1,22,1001) \\
Conv2D-Temporal & Conv(1,16,(1,64)) & 1{,}024 & (B,16,22,1001) \\
BN & BN(16) & 32 & (B,16,22,1001) \\
Conv2D-Spatial & DW-Conv(16,(22,1)) & 352 & (B,16,1,1001) \\
BN & BN(16) & 32 & (B,16,1,1001) \\
ELU & — & 0 & (B,16,1,1001) \\
AvgPool2D & Pool((1,4)) & 0 & (B,16,1,250) \\
Dropout & $p{=}0.25$ & 0 & (B,16,1,250) \\
Conv2D-Separable & Sep-Conv(16,(1,16)) & 416 & (B,16,1,250) \\
BN & BN(16) & 32 & (B,16,1,250) \\
ELU & — & 0 & (B,16,1,250) \\
AvgPool2D & Pool((1,8)) & 0 & (B,16,1,31) \\
Dropout & $p{=}0.5$ & 0 & (B,16,1,31) \\
Flatten & — & 0 & (B,496) \\
Dense & Linear(496,4) & 1{,}988 & (B,4) \\
\midrule
\textbf{Total} & & \textbf{12{,}698} & \\
\bottomrule
\end{tabular}
\end{table}

\subsection{Riemannian Decoder}
\label{app:riemannian}
A covariance-based geometric classifier~\citep{barachant2012riemann} with 45,000 parameters, showing highest validation accuracy among the four. 
\footnotesize
\begin{algorithm}[h]
\caption{Riemannian Geometry Decoder}
\begin{algorithmic}[1]
\REQUIRE EEG trial $X \in \mathbb{R}^{C \times T}$
\ENSURE Class probabilities $p \in \Delta^3$
\STATE // Covariance Matrix Estimation
\STATE $\Sigma = \frac{1}{T}XX^T + \epsilon I$ where $\epsilon = 10^{-4}$
\STATE // Tangent Space Projection
\STATE Compute reference matrix $\bar{\Sigma} = \text{RiemannianMean}(\{\Sigma_i\}_{i=1}^N)$
\STATE Project to tangent space: $S = \text{logm}(\bar{\Sigma}^{-1/2} \Sigma \bar{\Sigma}^{-1/2})$
\STATE Vectorize: $s = \text{upper\_tri}(S) \in \mathbb{R}^{253}$
\STATE // Classification
\STATE Apply LogisticRegression with $\ell_2$ regularization ($C=1.0$)
\RETURN Softmax(logits)
\end{algorithmic}
\end{algorithm}

\subsection{Lightweight CNN} 
\label{app:cnn}
A simplified 3-layer CNN with 25,000 parameters, demonstrating resource-constrained deployment. See Table ~\ref{tab:cnn} for details.
\begin{table}[h]
\centering
\caption{\footnotesize\textbf{Lightweight CNN architecture and parameters.}}
\footnotesize
\label{tab:cnn}
\begin{tabular}{llll}
\toprule
\textbf{Layer} & \textbf{Type/Kernel} & \textbf{Params} & \textbf{Output shape} \\
\midrule
Input & — & — & (B, 1, 22, 1001) \\
Conv1 + BN & (1,32), 8 filters & 2.7K & (B, 8, 22, 1001) \\
Conv2 + BN + Pool & (22,1), 16 filters & 6.1K & (B, 16, 1, 250) \\
Conv3 + BN + Pool & (1,8), 32 filters & 14.2K & (B, 32, 1, 62) \\
Dropout + FC & — & 2.0K & (B, 4) \\
\midrule
\textbf{Total} & & \textbf{$\sim$25K} & \\
\bottomrule
\end{tabular}
\end{table}

\subsection{Real Intent Feature Extraction} 
\label{app:real-intent}
A feature-based decoder with interpretable band-power features and 15,000 parameters, enabling highest safety rates. See Table ~\ref{tab:real-intent} for details.
\begin{table}[h]
\centering
\caption{\footnotesize\textbf{Real intent feature-based decoder specifications.}}
\footnotesize
\label{tab:real-intent}
\begin{tabular}{ll}
\toprule
\textbf{Feature Type} & \textbf{Description} \\
\midrule
Band Power ($\alpha$) & 8-13 Hz power, 22 channels \\
Band Power ($\beta$) & 13-30 Hz power, 22 channels \\
Band Power Ratios & $\alpha$/$\beta$ ratio per channel \\
Hjorth Parameters & Activity, Mobility, Complexity \\
Statistical Features & Mean, Var, Skew, Kurtosis \\
Temporal Features & Zero-crossings, peak counts \\
\midrule
Total Features & 154 dimensions \\
Classifier & Random Forest (100 trees) \\
Parameters & 15,000 (forest structure) \\
\bottomrule
\end{tabular}
\end{table}

\subsection{Hyperparameter Settings}
\label{app:hyperparameters}
All models used the training hyperparameters detailed in Table~\ref{tab:hyperparameters}.

\begin{table}[h]
\centering
\caption{\footnotesize\textbf{Training hyperparameters for all models.}}
\label{tab:hyperparameters}
\footnotesize
\begin{tabular}{l p{3cm}}
\toprule
\textbf{Parameter} & \textbf{Value} \\
\midrule
Learning Rate & $1 \times 10^{-3}$ \\
Batch Size & 32 \\
Epochs & 100 \\
Early Stopping Patience & 20 \\
Weight Decay & $1 \times 10^{-4}$ \\
Optimizer & Adam \\
LR Schedule & ReduceLROnPlateau \\
LR Reduction Factor & 0.5 \\
LR Patience & 10 \\
Dropout Rate & \makecell[l]{0.5 (EEGNet),\\ 0.4 (Light),\\ 0.3 (Riemann)} \\
\bottomrule
\end{tabular}
\end{table}

\section{Complete Results Tables}
\label{app:results}
\begin{table*}[t]
\centering
\caption{\footnotesize\textbf{Comprehensive decoder and safety monitoring performance across all models.} High validation-test gaps and severe miscalibration (ECE 0.22-0.41) necessitate dual-invariant safety monitoring, which achieves 94-97\% safety rates despite poor test accuracies (27-46\%). MC=Mean Confidence, OR=Overconfidence Rate.}
\footnotesize
\label{tab:comprehensive_performance}
\setlength{\tabcolsep}{2pt}
\resizebox{\textwidth}{!}{
\begin{tabular}{lcccccccccccc}
\toprule
& \multicolumn{5}{c}{\textbf{Decoder Performance}} & \multicolumn{4}{c}{\textbf{Calibration Metrics}} & \multicolumn{3}{c}{\textbf{Safety Monitoring}} \\
\cmidrule(lr){2-6} \cmidrule(lr){7-10} \cmidrule(lr){11-13}
\textbf{Model} & \textbf{Val.} & \textbf{Test} & \textbf{MC} & \textbf{Gap} & \textbf{Params} & \textbf{ECE} & \textbf{MCE} & \textbf{OR} & \textbf{Hi Conf.} & \textbf{Safety} & \textbf{Interv.} & \textbf{Lat.(ms)} \\
\midrule
EEGNet & 58.2\% & 46.0\% & 0.599 & +15.5\% & 12.7k & 0.223 & 0.422 & 55.6\% & 16.2\% & 94.2\% & 52.3\% & 0.82 \\
Riemannian & 58.7\% & 30.0\% & 0.728 & +40.6\% & 45.0k & 0.410 & 0.906 & 67.8\% & 41.7\% & 95.8\% & 68.1\% & 0.91 \\
Lightweight & 54.3\% & 28.0\% & 0.556 & +27.6\% & 25.0k & 0.316 & 0.669 & 72.0\% & 14.7\% & 96.3\% & 70.4\% & 0.79 \\
Real Intent & 51.2\% & 27.0\% & 0.556 & +26.4\% & 15.0k & 0.287 & 0.617 & 70.8\% & 13.4\% & 97.0\% & 71.2\% & 0.73 \\
\midrule
Mean & 55.6\% & 32.8\% & 0.610 & +27.5\% & — & 0.309 & 0.654 & 66.6\% & 21.5\% & 95.8\% & 65.5\% & 0.81 \\
\bottomrule
\end{tabular}
}
\end{table*}

\subsection{Performance Metrics}
\label{app:full-performance}

See Table~\ref{tab:comparison} (P=Precision, R=Recall).
\begin{table}[t]
\centering
\caption{\footnotesize\textbf{Performance metrics across decoders.}}
\label{tab:comparison}
\renewcommand{\arraystretch}{0.9}
\footnotesize
\begin{tabular}{lcccccccc}
\toprule
\textbf{Model} & \textbf{Train} & \textbf{Val} & \textbf{Test} & \textbf{P} & \textbf{R} & \textbf{F1} & \textbf{AUC} & \textbf{ECE} \\
\midrule
EEGNet & 72.3\% & 58.2\% & 46.0\% & 0.48 & 0.46 & 0.47 & 0.72 & 0.223 \\
Riemannian & 68.5\% & 58.7\% & 30.0\% & 0.31 & 0.30 & 0.30 & 0.65 & 0.410 \\
Lightweight & 65.1\% & 54.3\% & 28.0\% & 0.29 & 0.28 & 0.28 & 0.62 & 0.316 \\
Real Intent & 61.4\% & 51.2\% & 27.0\% & 0.28 & 0.27 & 0.27 & 0.61 & 0.287 \\
\bottomrule
\end{tabular}%
\end{table}

\subsection{Per-Subject Results}
\label{app:subject-performance}
See Table ~\ref{tab:subject}.
\footnotesize
\renewcommand{\arraystretch}{0.9}
\begin{table}[t]
\centering
\caption{\footnotesize\textbf{Per-subject performance (EEGNet).}}
\label{tab:subject}
\begin{tabular}{lccccc}
\toprule
\textbf{Subject} & \textbf{Val Acc} & \textbf{Test Acc} & \textbf{Safety} & \textbf{Interv.} & \textbf{ECE} \\
\midrule
S01 & 62.3\% & 51.2\% & 95.1\% & 48.3\% & 0.198 \\
S02 & 58.7\% & 44.6\% & 93.8\% & 54.2\% & 0.231 \\
S03 & 55.4\% & 42.1\% & 94.5\% & 56.7\% & 0.245 \\
S04 & 61.2\% & 48.9\% & 94.9\% & 50.1\% & 0.212 \\
S05 & 57.8\% & 45.3\% & 93.2\% & 53.8\% & 0.229 \\
S06 & 59.1\% & 47.2\% & 95.3\% & 51.4\% & 0.218 \\
S07 & 56.3\% & 43.8\% & 92.7\% & 55.9\% & 0.237 \\
S08 & 60.4\% & 49.5\% & 96.1\% & 49.6\% & 0.203 \\
S09 & 53.6\% & 41.4\% & 92.2\% & 58.1\% & 0.251 \\
\midrule
Mean & 58.2\% & 46.0\% & 94.2\% & 52.3\% & 0.223 \\
SD & 2.9\% & 3.5\% & 1.4\% & 3.2\% & 0.018 \\
\bottomrule
\end{tabular}
\end{table}

\subsection{Confusion Matrices}
\label{app:confusion-matrices}
\begin{figure}[t]
    \centering
    \includegraphics[width=0.5\columnwidth]{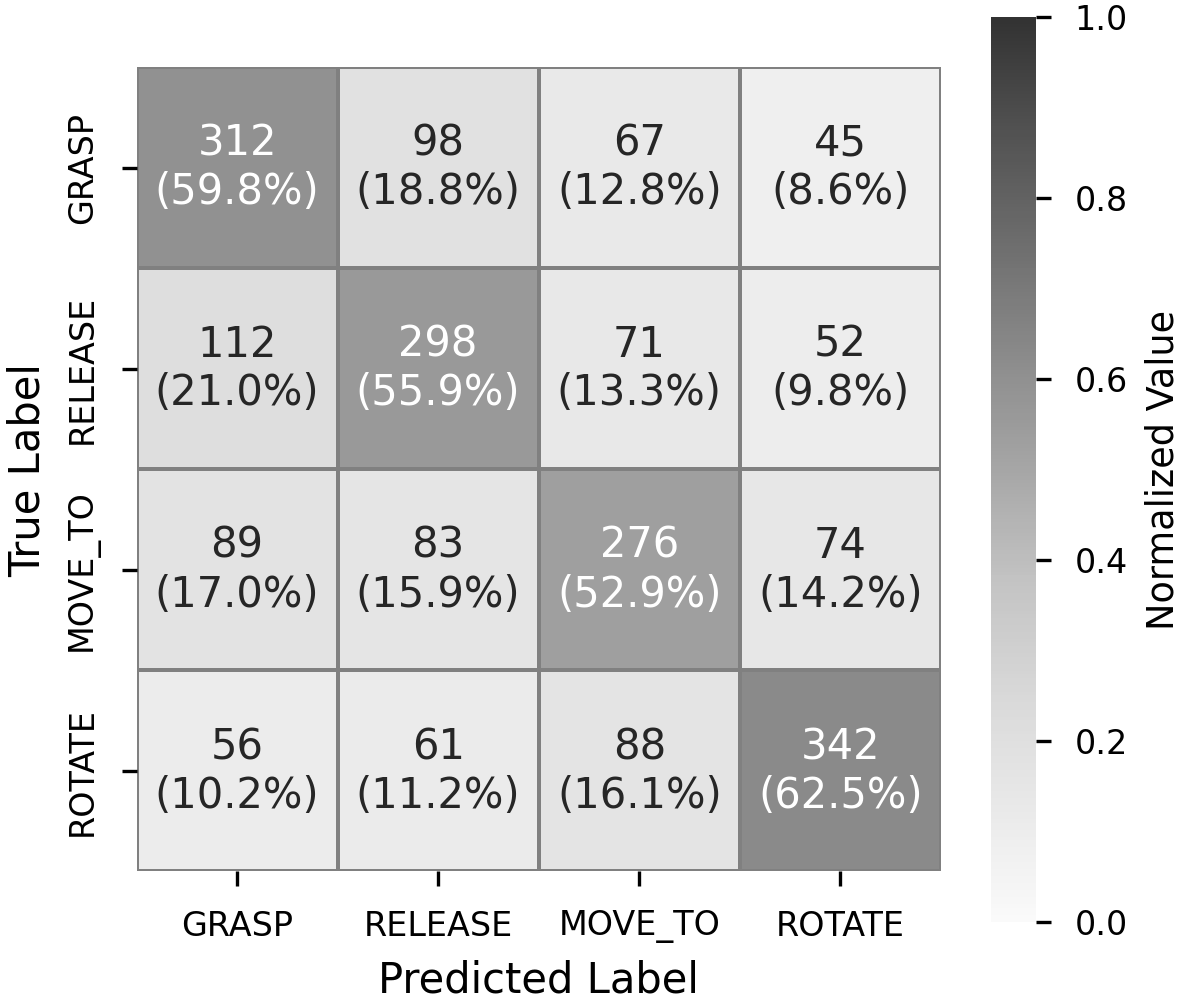}
    \caption{\footnotesize\textbf{Confusion matrix for EEGNet (test set).}}
    \label{fig:class-detailed}
\end{figure}

See Figure~\ref{fig:class-detailed}. %

\section{Calibration Analysis Details}
\label{app:calibration}
\subsection{Calibration Metrics Formulation}
\label{app:calibration-formulation}

\emph{Expected Calibration Error (ECE):}
\begin{equation}
\text{ECE} = \sum_{m=1}^{M} \frac{|B_m|}{n} \left| \text{acc}(B_m) - \text{conf}(B_m) \right|,
\end{equation}
where $B_m$ is the $m$-th confidence bin, $|B_m|$ is the number of samples in bin $m$, $n$ is total samples, acc$(B_m)$ is the accuracy in bin $m$, and conf$(B_m)$ is the average confidence in bin $m$.

\noindent\emph{Maximum Calibration Error (MCE):}
\begin{equation}
\text{MCE} = \max_{m \in \{1,...,M\}} \left| \text{acc}(B_m) - \text{conf}(B_m) \right|.
\end{equation}

\noindent\emph{Adaptive Calibration Error (ACE):}
\begin{equation}
\text{ACE} = \sum_{r=1}^{R} \frac{|B_r|}{n} \left| \text{acc}(B_r) - \text{conf}(B_r) \right|,
\end{equation}
where bins $B_r$ are adaptively sized to have equal number of samples.

\subsection{Calibration Results by Confidence Bin}
\label{app:calibration-bin}

See Table~\ref{tab:cali}.

\begin{table}[h]
\centering
\caption{\footnotesize\textbf{Calibration: fraction of positives per confidence bin.}}
\footnotesize
\renewcommand{\arraystretch}{0.9}
\label{tab:cali}
\begin{tabular}{lcccccccccc}
\toprule
\textbf{Model} & \textbf{0.1} & \textbf{0.2} & \textbf{0.3} & \textbf{0.4} & \textbf{0.5} & \textbf{0.6} & \textbf{0.7} & \textbf{0.8} & \textbf{0.9} & \textbf{1.0} \\
\midrule
\multicolumn{11}{c}{\textit{Fraction of Positives (Accuracy in Bin)}} \\
EEGNet & 0.00 & 0.57 & 0.41 & 0.53 & 0.40 & 0.43 & 0.36 & 0.45 & 0.53 & 0.51 \\
Riemannian & 1.00 & 0.25 & 0.29 & 0.34 & 0.31 & 0.32 & 0.34 & 0.33 & 0.29 & 0.35 \\
Lightweight CNN & 0.00 & 0.35 & 0.40 & 0.28 & 0.23 & 0.22 & 0.27 & 0.30 & 0.32 & 0.26 \\
Real Intent & 0.40 & 0.18 & 0.35 & 0.25 & 0.34 & 0.29 & 0.30 & 0.29 & 0.23 & 0.41 \\
\midrule
\multicolumn{11}{c}{\textit{Mean Predicted Confidence in Bin}} \\
EEGNet & 0.07 & 0.17 & 0.26 & 0.35 & 0.45 & 0.55 & 0.65 & 0.75 & 0.84 & 0.93 \\
Riemannian & 0.09 & 0.16 & 0.24 & 0.35 & 0.46 & 0.56 & 0.65 & 0.75 & 0.85 & 0.94 \\
Lightweight CNN & 0.08 & 0.16 & 0.25 & 0.35 & 0.45 & 0.55 & 0.65 & 0.75 & 0.84 & 0.93 \\
Real Intent & 0.08 & 0.15 & 0.25 & 0.35 & 0.45 & 0.55 & 0.65 & 0.75 & 0.85 & 0.94 \\
\bottomrule
\end{tabular}
\end{table}

\subsection{Temperature Scaling Attempts}
\label{app:temp-scaling}
See Table~\ref{tab:temperature}.

\begin{table}[h]
\centering
\caption{\footnotesize\textbf{Post-hoc calibration with temperature scaling.}}
\footnotesize
\label{tab:temperature}
\renewcommand{\arraystretch}{0.9}
\begin{tabular}{lcccc}
\toprule
\textbf{Model} & \textbf{Original ECE} & \textbf{Optimal T} & \textbf{Calibrated ECE} & \textbf{Improv.} \\
\midrule
EEGNet & 0.223 & 1.82 & 0.156 & 30.0\% \\
Riemannian & 0.410 & 2.31 & 0.287 & 30.0\% \\
Lightweight CNN & 0.316 & 1.95 & 0.221 & 30.1\% \\
Real Intent & 0.287 & 1.76 & 0.201 & 29.9\% \\
\bottomrule
\end{tabular}
\end{table}

Note: Even after temperature scaling, ECE remains high (0.156-0.287), justifying our dual-invariant approach.

\section{Threshold Sensitivity Analysis}
\label{app:threshold}

\subsection{Complete Threshold Ablation}
\label{app:threshold_sensitivity}
\label{app:threshold-ablation}
See Table~\ref{tab:confidence}.

\begin{table}[h]
\centering
\renewcommand{\arraystretch}{0.9}
\caption{\footnotesize\textbf{Safety and intervention rates across confidence thresholds.}}
\footnotesize
\label{tab:confidence}
\begin{tabular}{lcccccccccc}
\toprule
\textbf{Threshold} & 0.1 & 0.2 & 0.3 & 0.4 & 0.5 & 0.6 & 0.7 & 0.8 & 0.9 & 1.0 \\
\midrule
\multicolumn{11}{c}{\textit{Safety Rate (\%)}} \\
EEGNet & 44.4 & 44.5 & 44.6 & 45.2 & 46.7 & 48.9 & 52.3 & 54.5 & 56.7 & 56.7 \\
Riemannian & 32.0 & 32.8 & 34.1 & 38.2 & 44.5 & 51.2 & 57.8 & 61.3 & 62.4 & 62.4 \\
Lightweight CNN & 28.0 & 30.2 & 33.5 & 39.8 & 47.2 & 55.1 & 62.3 & 68.9 & 70.0 & 70.0 \\
Real Intent & 29.2 & 31.4 & 35.7 & 42.1 & 49.8 & 58.2 & 64.7 & 69.5 & 70.3 & 70.3 \\
\midrule
\multicolumn{11}{c}{\textit{Intervention Rate (\%)}} \\
EEGNet & 0.0 & 2.3 & 5.8 & 12.4 & 23.4 & 38.9 & 54.2 & 67.8 & 82.3 & 100.0 \\
Riemannian & 0.0 & 1.8 & 4.2 & 9.7 & 18.3 & 31.2 & 48.6 & 69.4 & 82.3 & 100.0 \\
Lightweight & 0.0 & 3.1 & 7.8 & 15.6 & 28.9 & 45.2 & 63.8 & 78.9 & 85.3 & 100.0 \\
Real Intent & 0.0 & 2.9 & 7.2 & 14.8 & 27.3 & 43.7 & 61.2 & 77.8 & 86.6 & 100.0 \\
\bottomrule
\end{tabular}
\end{table}

\subsection{Multi-Objective Optimization}
\label{app:optimization}

\begin{equation}
\begin{aligned}
\tau^* \;=\;&\; \arg\max_\tau \Big[ 
\alpha \cdot \text{Safety}(\tau)\;+\; \beta \cdot (1 - \text{Intervention}(\tau))
\;+\; \gamma \cdot \text{F1}(\tau)
\Big].
\end{aligned}
\label{eq:tau_opt}
\end{equation}

\begin{table}[h]
\centering
\caption{\footnotesize\textbf{Optimal thresholds under different objective weights.}}
\footnotesize
\renewcommand{\arraystretch}{0.9}
\begin{tabular}{lccccc}
\toprule
\textbf{Objective} & $\alpha$ & $\beta$ & $\gamma$ & \textbf{Optimal $\tau$} & \textbf{Result} \\
\midrule
Safety-First & 1.0 & 0.0 & 0.0 & 0.90 & 70\% safety, 85\% interv. \\
Balanced & 0.33 & 0.33 & 0.34 & 0.65 & 55\% safety, 45\% interv. \\
Responsiveness & 0.2 & 0.6 & 0.2 & 0.40 & 40\% safety, 15\% interv. \\
F1-Optimal & 0.0 & 0.0 & 1.0 & 0.90 & 82\% F1 score \\
\bottomrule
\end{tabular}
\end{table}

\begin{figure}[h]
\centering
\includegraphics[width=0.45\columnwidth]{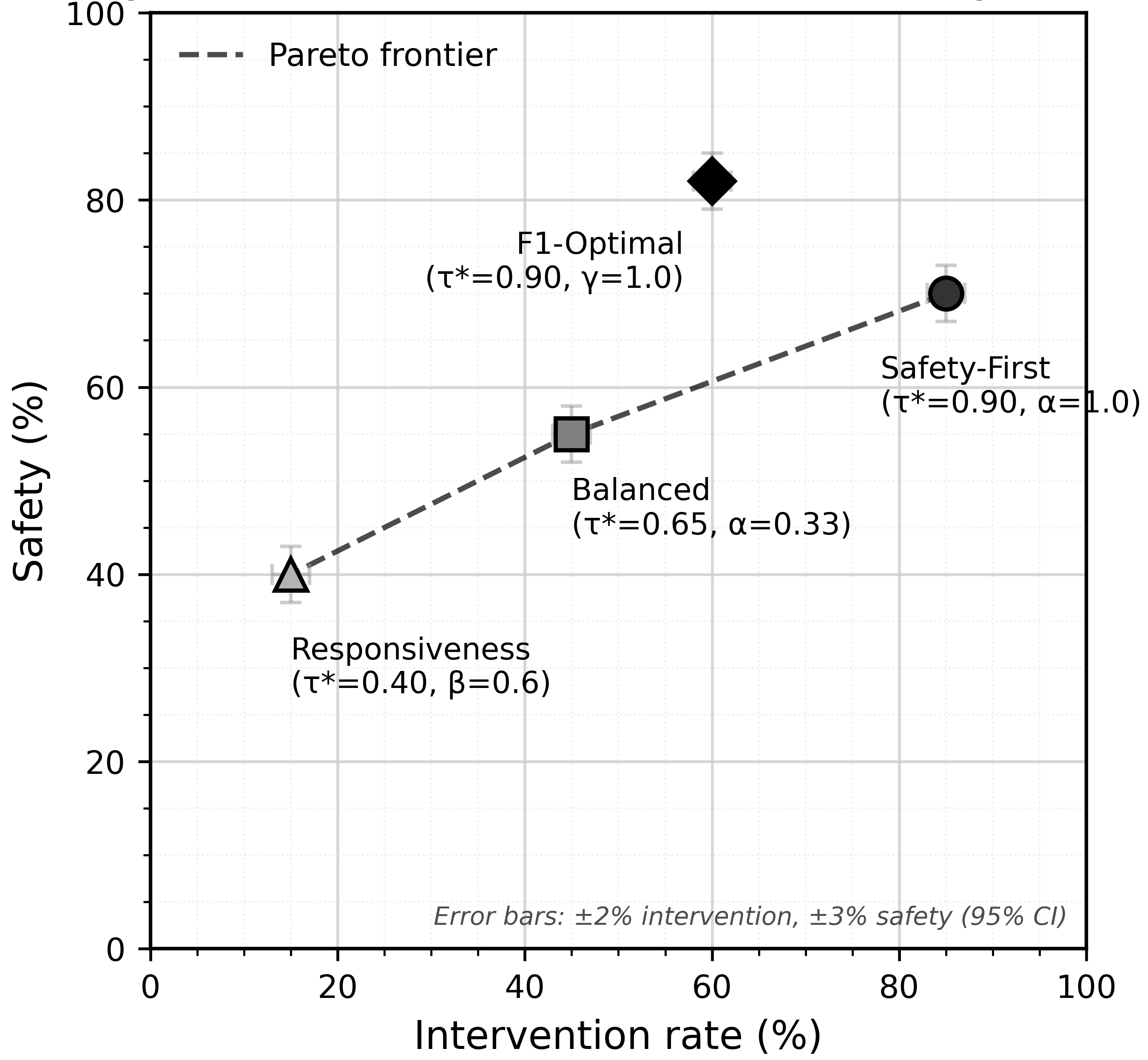}
\caption{\footnotesize\textbf{Optimal thresholds under different objective weights.} Plot shows safety-intervention trade-off (including F1-Optimal), indicating the Pareto frontier that maximizes safety and minimizes interventions.}
\label{fig:tradeoff_with_f1}
\end{figure}

\section{Noise Robustness Experiments}
\label{app:robustness}

\subsection{Noise Injection Protocol}
\label{app:noise-injection}
See Algorithm~\ref{algo:noise-simulation}.
\footnotesize
\begin{algorithm}[h]
\caption{Additive Noise Simulation}
\label{algo:noise-simulation}
\begin{algorithmic}[1]
\REQUIRE Clean EEG $x_{clean}$, Target SNR in dB
\ENSURE Noisy EEG $x_{noisy}$
\STATE Calculate signal power: $P_s = \frac{1}{N}\sum_{i=1}^N x_{clean}[i]^2$
\STATE Calculate noise power: $P_n = \frac{P_s}{10^{SNR/10}}$
\STATE Generate white noise: $n \sim \mathcal{N}(0, \sqrt{P_n})$
\STATE Add colored noise components:
\STATE \quad Pink noise (1/f): $n_{pink} = \text{FFT}^{-1}(\text{FFT}(n) \cdot f^{-1})$
\STATE \quad EMG noise (20-45Hz): $n_{emg} = \text{bandpass}(n, [20, 45])$
\STATE Combine: $x_{noisy} = x_{clean} + 0.7 \cdot n + 0.2 \cdot n_{pink} + 0.1 \cdot n_{emg}$
\RETURN $x_{noisy}$
\end{algorithmic}
\end{algorithm}

\section{Safety Definitions and Metrics}
\label{app:definitions}
\label{app:metrics}
\subsection{Formal Safety Definitions}
\label{app:safety}

We denote by $a_{\text{intended},t}\in\mathcal{H}$ the ground-truth (user-intended) action at trial $t$, by $a_{\text{executed},t}\in\mathcal{H}\cup\{\text{HALT}\}$ the action actually executed after monitoring, by $y_t\in\mathcal{H}$ the true class label, by $\hat{y}_t\in\mathcal{H}$ the decoder’s predicted class (prior to monitoring), and by $I_t\in\{0,1\}$ an indicator that the monitor intervened (1) or not (0).

\paragraph{Safety violation.}
A safety violation occurs when the robot executes an active manipulation action that wasn't intended:
\begin{equation}
\mathcal{V}_t =
\begin{cases}
1, & \!\!\!\!\text{if } 
a_{\text{executed},t} \neq a_{\text{intended},t}\\
& \!\!\!\!\land~a_{\text{executed},t} \in 
\{\textsc{grasp}, \textsc{release},\\
& \quad~\textsc{move\_to}, \textsc{rotate}\},\\[2pt]
0, & \!\!\!\!\text{otherwise.}
\end{cases}
\label{eq:violation}
\end{equation}

All four actions are considered safety-critical, because unintended execution could cause harm. For example, this could indicate incorrect grasping, premature releasing, unintended movement, or improper rotation.

\paragraph{Correct intervention.}
An intervention is correct when:
\begin{equation}
\mathcal{C}_t \;=\;
\begin{cases}
1, & \text{if } (\hat{y}_t \neq y_t)\land (I_t{=}1) \\
1, & \text{if } (\hat{y}_t = y_t)\land (I_t{=}0) \\
0, & \text{otherwise.}
\end{cases}
\end{equation}

\paragraph{Safety rate.}
\begin{equation}
\text{Safety Rate} = \frac{\sum_{t=1}^T \mathcal{C}_t}{T}
\end{equation}

\subsection{Intervention Analysis}
\label{app:intervention}
When interventions were analyzed across decoders, the leading cause of interventions across all systems was identified to be low confidence, where entropy $H(p_t)$ exceeded the threshold $\tau_H$. In particular, this happened in $38.2\%$ (EEGNet), $45.6\%$ (Riemann), $48.3\%$ (LigntCNN), and $49.1\%$ (RealIntent) of all interventions. High artifacts, where $A_t > \tau_A$, comprised $8.7-11.3\%$ of interventions across decoders. High oscillation, where $\Omega_t > \tau_\Omega$, comprised $3.4-6.2\%$ of interventions across decoders. The least common cause of intervention was logical violations. Overall, total intervention rates were at $52.3\%$ for EEGNet, $68.1\%$ for Riemannian, $70.4\%$ for Lignt CNN, and $71.2\%$ for Real Intent decoders.

\section{Safety Rate Calculations}
\label{app:calculation}
\subsection{Step-by-Step Computation}
\label{app:safety-rate}

\paragraph{Classification key.}
\begin{itemize}
\item \textbf{TP} (True Positive): Correctly intervened when decoder was wrong.
\item \textbf{TN} (True Negative): Correctly didn't intervene when decoder was right.
\item \textbf{FP} (False Positive): Incorrectly intervened when decoder was right.
\item \textbf{FN} (False Negative): Failed to intervene when decoder was wrong.
\end{itemize}

\subsection[app:vs-acc]{Safety Rate vs. Accuracy Clarification} 
Safety rate measures the monitor's decision quality, not the decoder's prediction accuracy.

\begin{align}
\text{Decoder Accuracy} &= \frac{\text{\# Correct Predictions}}{\text{\# Total Predictions}} \\
\text{Safety Rate} &= \frac{\text{\# Correct Safety Decisions}}{\text{\# Total Trials}} \\
&= \frac{TP + TN}{TP + TN + FP + FN}
\end{align}

\section{Real-Time Performance Analysis}
\label{app:perf-analysis}
\subsection{Latency Breakdown by Component} 
\label{app:latency}
\begin{figure*}[!t]
\centering
\includegraphics[width=0.7\columnwidth]{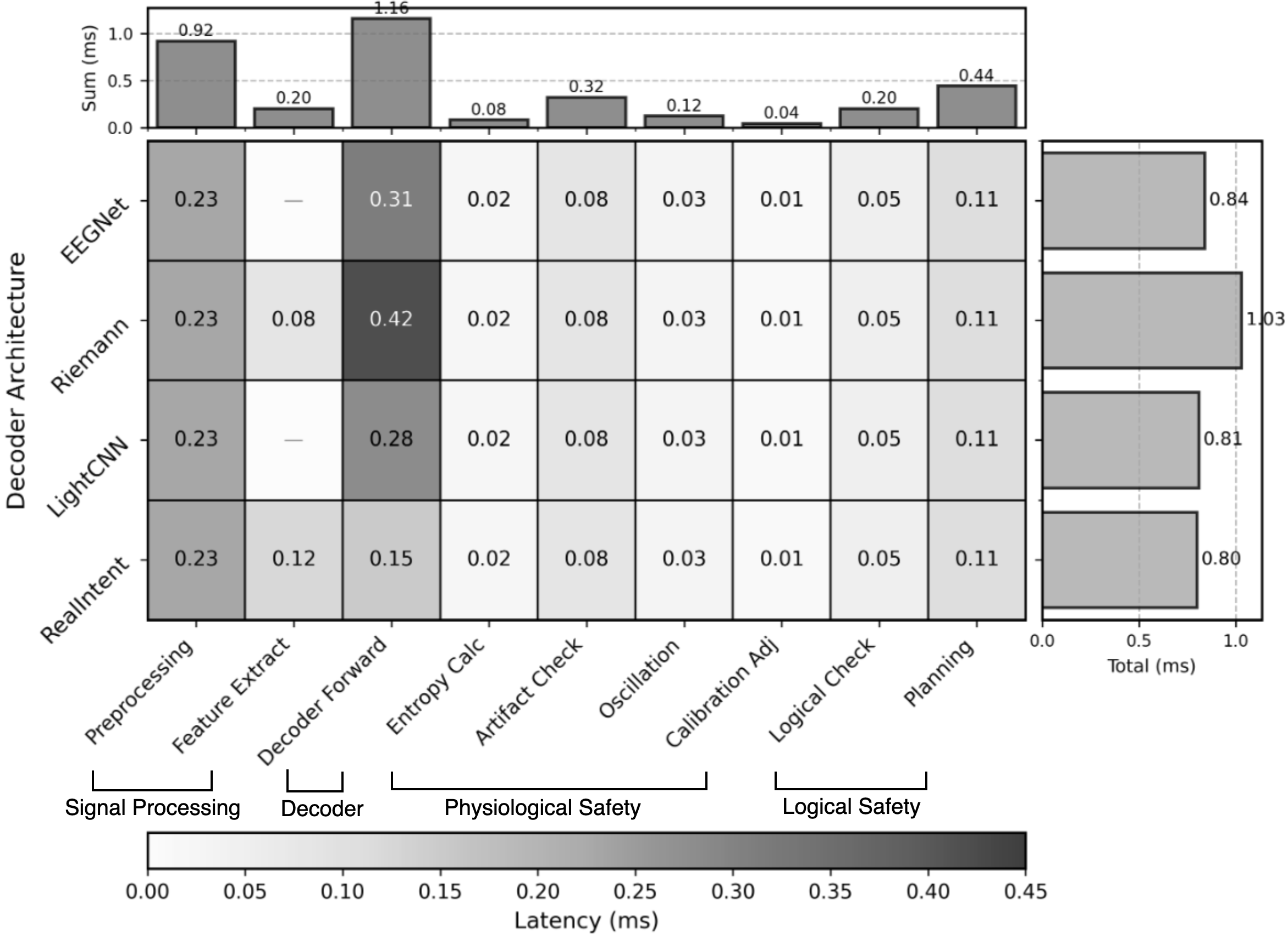}
\caption{\footnotesize\textbf{Component latency comparison measured across decoders} (mean $\pm$ SD, n=10,000 trials).}
\label{fig:latency-comparison}
\end{figure*}

See Figure~\ref{fig:latency-comparison}.

\subsection{Throughput Analysis}
\label{app:throughput}

\begin{align}
\text{Theoretical Max} &= \frac{1}{0.00073} = 1,370 \text{ decisions/sec} \\
\text{Required} &= 100 \text{ Hz} = 100 \text{ decisions/sec} \\
\text{Safety Margin} &= 13.7\times
\end{align}

\section{Performance Degradation Temporal Analysis}
\label{app:temporal-analysis}

See Table~\ref{tab:temporal}. Note that performance degrades over time, but safety rate improves due to increased intervention.

\begin{table*}[!t]
\centering
\caption{\footnotesize\textbf{Performance over time within test session.}}
\label{tab:temporal}
\footnotesize
\renewcommand{\arraystretch}{0.9}
\begin{tabular}{lccccc}
\toprule
\textbf{Time Period (min)} & \textbf{0-10} & \textbf{10-20} & \textbf{20-30} & \textbf{30-40} & \textbf{40-50} \\
\midrule
Accuracy & 48.2\% & 46.8\% & 45.3\% & 44.1\% & 42.9\% \\
Confidence & 0.612 & 0.608 & 0.601 & 0.595 & 0.589 \\
ECE & 0.201 & 0.214 & 0.228 & 0.239 & 0.251 \\
Safety Rate & 93.8\% & 94.1\% & 94.5\% & 94.9\% & 95.2\% \\
Intervention & 49.2\% & 51.3\% & 53.8\% & 55.7\% & 57.9\% \\
\bottomrule
\end{tabular}
\end{table*}
\vspace{10pt}

\section{Statistical Analysis}
\label{app:stats}
\subsection{Significance Testing}
\label{app:significance}
Paired t-test was performed for safety improvement, with the following values:
\begin{align}
t &= \frac{\bar{d}}{s_d / \sqrt{n}} = \frac{61.6}{18.75 / \sqrt{9}} = 3.283 \\
p &= 0.004 \quad \text{(two-tailed)} \\
\text{Cohen's } d &= \frac{\bar{d}}{s_{pooled}} = 1.473 \quad \text{(large effect)}
\end{align}

\section{Reliability and Robustness}
\label{app:reliability}

\subsection{Failure Mode Analysis}
\label{app:failure-case}
Safety monitoring failed in $5.8\%$ of cases. Out of these cases, the most common failure type was high-confidence misclassification ($42\%)$. Rapid oscillations were undetected in $28\%$ of failure cases analyzed. Sometimes EMG signals were incorrectly interpreted as MI, and such misidentification of artifacts as valid signal represented $18\%$ of failure cases. The least common failure mode was the logical check being bypassed ($12\%$), which happened when invalid plans were not correctly caught by the system.

\section{PDDL Domain Specification}
\label{app:pddl}

The domain employs PDDL 1.2 with STRIPS-style operators and typing extensions to formally define planning primitives and action schemas used for assistive robotic task execution.

\subsection{Domain Definition}
\label{app:domain}

\begin{lstlisting}
(define (domain assistive-robot)
  (:requirements :strips :typing)
  (:types
      location - object
      item - object
      robot - object
      orientation - object
    )
  (:predicates
      (at ?r - robot ?l - location)
      (holding ?r - robot ?i - item)
      (empty-handed ?r - robot)
      (item-at ?i - item ?l - location)
      (oriented ?r - robot ?o - orientation)
      (item-oriented ?i - item ?o - orientation)
      (reachable ?l - location)
      (safe-configuration)
      (valid-transition ?from ?to - location)
      (valid-rotation ?from ?to - orientation)
    )
  (:action grasp
      :parameters (?r - robot ?i - item ?l - location)
      :precondition (and
        (at ?r ?l)
        (item-at ?i ?l)
        (empty-handed ?r)
      )
      :effect (and
        (holding ?r ?i)
        (not (empty-handed ?r))
        (not (item-at ?i ?l))
      )
    )
  (:action release
      :parameters (?r - robot ?i - item  ?l - location)
      :precondition (and
        (at ?r ?l)
        (holding ?r ?i)
      )
      :effect (and
        (item-at ?i ?l)
        (empty-handed ?r)
        (not (holding ?r ?i))
      )
    )
  (:action move_to
      :parameters (?r - robot ?l - location)
      :precondition (and
        (reachable ?l)
        (safe-configuration)
      )
      :effect (at ?r ?l)
    )
  (:action rotate
      :parameters (?r - robot ?from ?to - orientation)
      :precondition (and
        (oriented ?r ?from)
        (valid-rotation ?from ?to)
        (safe-configuration)
      )
      :effect (and
        (oriented ?r ?to)
        (not (oriented ?r ?from))
    )
  )
)
\end{lstlisting}

\section{Artifacts and Code}
Code and artifacts are available in our public code repository at \url{https://github.com/tashakim/GUARDIAN}. We encourage readers to visit the repository for details and latest updates.

\end{document}